# RCL-Mamba: A Dual-domain State Space Model for Measurement-oriented Image Restoration in Rotational Sparse-View Scanning Computed Laminography

Xuyang Duan [a, b], Genyuan Zhang [a, b], Zhenjiang Dong [a, b], Chuandong Tan [a, b], Zihao Wang [a, b], Junyao Wang [a, b], Fenglin Liu [a, b, *]

[a] *Key Laboratory of Optoelectronic Technology and Systems, Ministry of Education, Chongqing University, Chongqing, 400044, China*

[b] *Engineering Research Center of Industrial CT Nondestructive Testing, Ministry of Education, Chongqing University, Chongqing, 400044, China*

** Author to whom any correspondence should be addressed.*

*E-mail: liufl@cqu.edu.cn*

**Abstract:** Rotational Scanning Computed Laminography (RCL) is widely utilized for the Non-Destructive Testing (NDT) of large planar components. However, to facilitate rapid inspection, continuous sparse-view scanning is often employed, where the angular integration effect during exposure induces rotational blur in the projection domain. Furthermore, the data incompleteness inherent in sparse sampling manifests as sparse artifacts in the reconstructed image domain. To address these cross-domain degradations, this paper proposes RCL-Mamba, a measurement-oriented dual-domain State Space Model (SSM)-based image restoration network. The framework adopts a cascaded joint processing strategy: it first corrects the rotational blur in the projection domain and subsequently suppresses the sparse artifacts in the image domain. Additionally, we design a Mamba-CNN dual-branch module to adaptively balance large-scale blur correction with local detail recovery. Evaluations on both simulated datasets and real-world Printed Circuit Board (PCB) scans demonstrate that RCL-Mamba outperforms existing baselines in blur removal, artifact suppression, and structural preservation. Line-profile-based structural measurement further verifies that the proposed method better preserves via/pad boundaries and slender trace profiles. Crucially, by reducing the required scanning views from 512 to 64, our method enhances inspection efficiency by approximately 8-fold without compromising reconstruction quality, offering a robust measurement-oriented restoration solution for high-throughput RCL inspection with improved structural measurement fidelity.

## 1. Introduction

Rotational Scanning Computed Laminography (RCL) [1] is widely utilized in Non-Destructive Testing (NDT) applications, particularly for electronic packaging and printed circuit board (PCB) inspection[2-3]. However, to satisfy the practical demands of high-throughput inspection, RCL systems typically employ continuous sparse-view scanning[4]. While efficient, this acquisition mode inevitably induces complex cross-domain composite degradation. Specifically, during a single exposure interval, the continuous rotation of the inspected object forces the detector to record the angular integral of multiple instantaneous projections, thereby generating severe rotational blur in the projection domain. Furthermore, the inherent data incompleteness of sparse-view sampling severely compromises reconstruction fidelity, manifesting as pronounced sparse artifacts and structural distortions in the image domain[5-6]. Consequently, mitigating these coupled cross-domain degradations is an imperative prerequisite for reliable structural measurement and precise identification of minute defects in practical high-throughput RCL inspection[7].

In recent years, deep learning methods have substantially advanced image deblurring tasks, with existing paradigms primarily categorized into Convolutional Neural Network (CNN)-based and Transformer-based approaches[8]. CNN-based methods [9-16] generally enhance restoration performance via multi-scale modeling, feature fusion, and lightweight designs. For instance, NAFNet [14] achieves efficient restoration through a simplified architecture, while MLWNet-B [15] bolsters context modeling by incorporating multi-level wavelet transforms. Specifically within RCL scenarios, MAFusNet [17] endeavors to mitigate rotational blur and artifacts by leveraging multi-angle projection feature fusion. Nevertheless, inherently constrained by local receptive fields and the assumption of spatial stationarity, these convolution-dependent methods struggle to model the large-scale, non-uniform rotational blur that expands with the rotation radius in RCL. Consequently, they exhibit evident limitations when recovering from severe blur trajectories in peripheral regions. To overcome these local constraints and effectively capture long-range dependencies, Transformers have been introduced into image restoration. Representative frameworks include Restormer [18], approaches utilizing stripe and window attention mechanisms [19, 20], FFTformer [21], which integrates frequency-domain interactions, and MDT [22], based on motion vector decomposition. Although these architectures excel at global information representation, they inherently impose prohibitive memory consumption and computational overhead when processing high-resolution industrial RCL projection data.

State Space Models (SSMs) [23, 24], particularly Mamba [25], provide a novel perspective for RCL image restoration. Unlike CNNs, Mamba circumvents the restrictions of local receptive fields, making it exceptionally adept at characterizing long-range degradation trajectories. Furthermore, compared to Transformers, Mamba achieves long-range dependency modeling with linear complexity via its selective scan mechanism (S6), thereby offering a distinct computational advantage when processing high-resolution projection data. The rotational blur induced during RCL continuous scanning exhibits severe large-scale, highly anisotropic, and spatially variant degradation intensities. These complex physical characteristics align seamlessly with Mamba’s inherent proficiency in capturing long-range, non-uniform dependencies. Consequently, several recent Visual State Space Models (VSSMs) [26-32], such as EAMamba [29] and MaIR [32], have demonstrated promising performance in image restoration tasks, successfully enhancing their adaptability to two-dimensional (2D) structures through customized spatial scanning strategies or geometric transformations.

Despite their promising performance, existing VSSMs primarily target general degradation in natural images, rendering them inadequate for the complex composite distortions induced under RCL continuous sparse-view scanning. Moreover, prevailing methods predominantly adhere to a single-domain restoration paradigm, inherently neglecting the physical correlation between projection-domain rotational blur and image-domain sparse artifacts. From the perspective of RCL imaging physics, this degradation exhibits distinct stage-wise and cross-domain characteristics: rotational blur originates during the projection acquisition stage and subsequently propagates through the reconstruction process, manifesting as severe sparse artifacts and structural distortions in the image domain. Therefore, to effectively mitigate cross-domain error accumulation, it is imperative to execute targeted processing across both domains in strict accordance with the physical degradation sequence. Furthermore, the composite degradation in RCL encompasses not only large-scale rotational blur but also the severe loss of high-frequency local textures and edge information. Although SSMs are highly advantageous for modeling long-range global dependencies, their capacity to reconstruct localized micro-structures requires further enhancement [31, 32]. Consequently, synergizing dual-domain joint processing with adaptive global-local feature modeling is essential for achieving high-fidelity image restoration in RCL. More importantly, for measurement-oriented RCL inspection, restoration performance should not be assessed only by global image quality metrics, but also by the preservation of local structural profiles and boundary consistency of PCB microstructures[33].

To bridge these technical gaps and align with the physical imaging sequence, this paper proposes RCL-Mamba, a novel image restoration framework based on joint projection-image dual-domain processing. This framework decomposes the complex restoration task into two cascaded stages. First, the rotational blur induced by continuous exposure is corrected in the projection domain, fundamentally recovering clear projections that approximate ideal static scanning conditions. Subsequently, following Feldkamp-Davis-Kress (FDK) reconstruction, the residual sparse artifacts and local structural distortions are suppressed in the image domain. Furthermore, to adaptively balance large-scale degradation modeling with microstructure preservation, a Mamba-CNN dual-branch module is designed. Specifically, the Mamba branch exploits the selective scan mechanism of the SSM to efficiently model global, non-uniform blur trajectories, while the CNN branch utilizes local convolutions to supplement high-frequency textures and edge details. A dynamic gated fusion mechanism is then employed to achieve adaptive synergy between these global and local representations. By synergizing the dual-domain cascaded strategy with the dual-branch modeling mechanism, RCL-Mamba rigorously conforms to the physical imaging process of RCL, achieving superior restoration performance and robustness under diverse data conditions. The main contributions of this paper are summarized as follows:

(1) State Space Models (SSMs) are introduced to the RCL deblurring task for the first time, achieving targeted modeling of complex rotational blur with a global receptive field and linear computational complexity.

(2) A projection-image dual-domain cascaded restoration strategy is proposed. By correcting rotational blur in the projection domain and suppressing sparse artifacts in the image domain, the framework strictly aligns with the physical degradation sequence, substantially enhancing restoration efficacy and generalizability.

(3) A Mamba-CNN dual-branch module integrated with a dynamic gated fusion mechanism is designed. This architecture realizes the adaptive synergy between global degradation modeling and local detail representation, effectively balancing large-scale blur correction with high-frequency detail preservation.

(4) Evaluations on simulated and real-world PCB datasets demonstrate that RCL-Mamba significantly outperforms existing methods in terms of reconstruction quality and structural measurement fidelity. By reducing the required scanning views from 512 to 64, the proposed method maintains high-fidelity reconstruction and local profile consistency while increasing single-station inspection efficiency by approximately 8-fold.

## 2. Related Work

The continuous-time SSM[23, 24] maps an input signal $x(t)$ to an output response $y(t)$ through a hidden state $h(t)$. This physical system can typically be described by a set of first-order ordinary differential equations:

$$h'(t) = \mathbf{A}h(t) + \mathbf{B}x(t), \tag{1}$$

$$y(t) = \mathbf{C}h(t) + \mathbf{D}x(t), \tag{2}$$

where $\mathbf{A}$, $\mathbf{B}$, $\mathbf{C}$, and $\mathbf{D}$ represent the state transition matrix, input matrix, output matrix, and direct transmission matrix, respectively. For discrete signals, SSMs generally employ the Zero-Order Hold (ZOH) discretization method to convert the continuous form into a discrete recursive form:

$$h_t = \overline{\mathbf{A}}h_{t-1} + \overline{\mathbf{B}}x_t, \tag{3}$$

$$y_t = \mathbf{C}h_t + \mathbf{D}x_t \tag{4}$$

$$\overline{\mathbf{A}} = e^{\Delta A}, \quad \overline{\mathbf{B}} = (\Delta\mathbf{A})^{-1}\left(e^{\Delta\mathbf{A}} - I\right)\cdot\Delta\mathbf{B}, \tag{5}$$

where $\overline{\mathbf{A}}$, $\overline{\mathbf{B}}$, etc., denote the discretized system parameters. Based on this discretized framework, the Mamba model introduces a Selective Scan Mechanism (S6)[25] equipped with dynamic weights. This mechanism achieves input-dependent dynamic modeling while strictly maintaining linear computational complexity.

However, Mamba is inherently tailored for one-dimensional (1D) sequence processing. Directly flattening two-dimensional (2D) image features would inevitably disrupt their original spatial adjacency relationships. To address this structural incompatibility, VSSMs[26-32] typically introduce customized 2D scanning strategies to map spatially topological image features into ordered sets of 1D sequences. This transformation facilitates the global state recurrence of 2D features without compromising their intrinsic spatial structures. Given the distinct linear complexity advantage of VSSMs in capturing long-range, non-local dependencies, VSSMs are introduced into the proposed network architecture in this study to effectively tackle the large-scale, spatially variant degradation characteristics inherent in RCL scanning.

## 3. Methodology

### 3.1 RCL Rotational Blur

Under ideal static scanning conditions, the detector records the instantaneous projection of the object at each discrete angular position. Assuming the ideal projection is denoted as $P_{ideal}$ and the reconstruction operator is defined as $\mathcal{R}^T$, the ideal ground-truth slice image, $I_{gt}$ can be expressed as:

$$I_{gt} = \mathcal{R}^T P_{ideal} \tag{6}$$

However, under practical continuous sparse-view scanning, the inspected object rotates continuously during a single exposure interval. Consequently, the detector records the integral average of multiple instantaneous projections $P_\theta$ over a limited exposure angular range. Therefore, the actually acquired blurry projection, denoted as $P_{blur}(u,v)$ can be formulated as:

$$P_{blur}(u,v) = \frac{1}{\Delta\theta}\int_{\theta_0}^{\theta_0+\Delta\theta} P_\theta(u,v)\,d\theta \approx \frac{1}{K}\sum_{i=1}^{K} P_{\theta_i}(u,v), \tag{7}$$

where $(u,v)$ denotes the detector coordinates, $\theta_0$ is the initial exposure angle, $\Delta\theta$ represents the angular range corresponding to a single exposure, and K is the number of instantaneous projection frames utilized for the discrete approximation of the continuous integral. Equation (7) mathematically indicates that the rotational blur in the RCL projection domain fundamentally originates from the angular integration effect during continuous exposure. This physical generation mechanism is illustrated in Fig. 1.

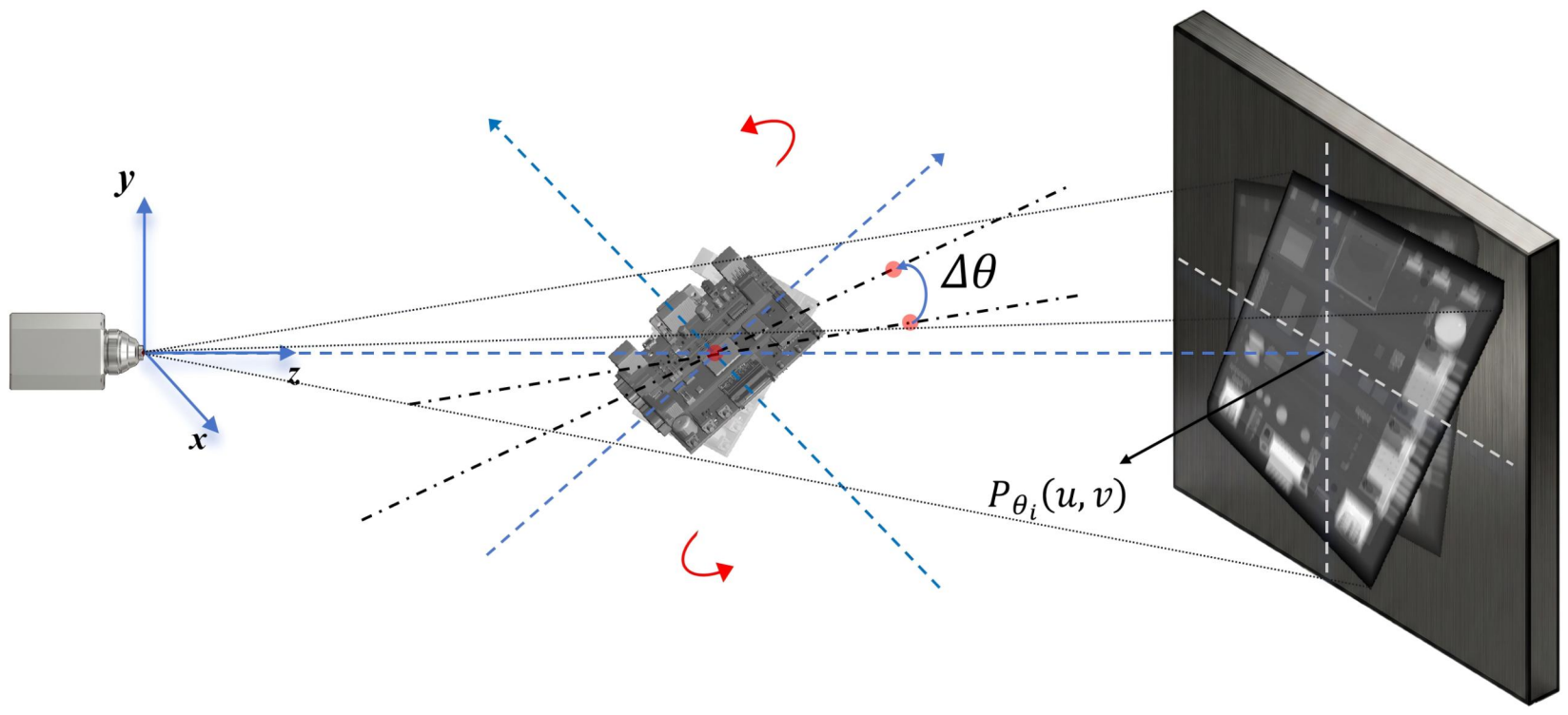

**Fig. 1.** Generation mechanism of rotational blur in RCL continuous scanning. During a single exposure interval, the inspected object rotates continuously across an angular range $\Delta\theta$. Consequently, the projection data recorded by the detector becomes the integral of multiple instantaneous projections $P_{\theta_i}(u,v)$ within that range, thereby inducing severe rotational blur in the projection domain.

If the blurry projection $P_{blur}(u,v)$ is directly reconstructed using the Feldkamp-Davis-Kress (FDK) algorithm [34] via the operator $\mathcal{R}^T$, the rotational blur in the projection domain inherently propagates into the image domain during the back-projection process. As indicated by Equation (6), this degradation exhibits distinct cross-domain coupling: the rotational blur originates from the angular integration effect in the projection domain and subsequently leads to severe structural blurring and detail distortion in the reconstructed slices.The manifestation of rotational blur across both the projection and image domains is visually compared in Fig. 2.

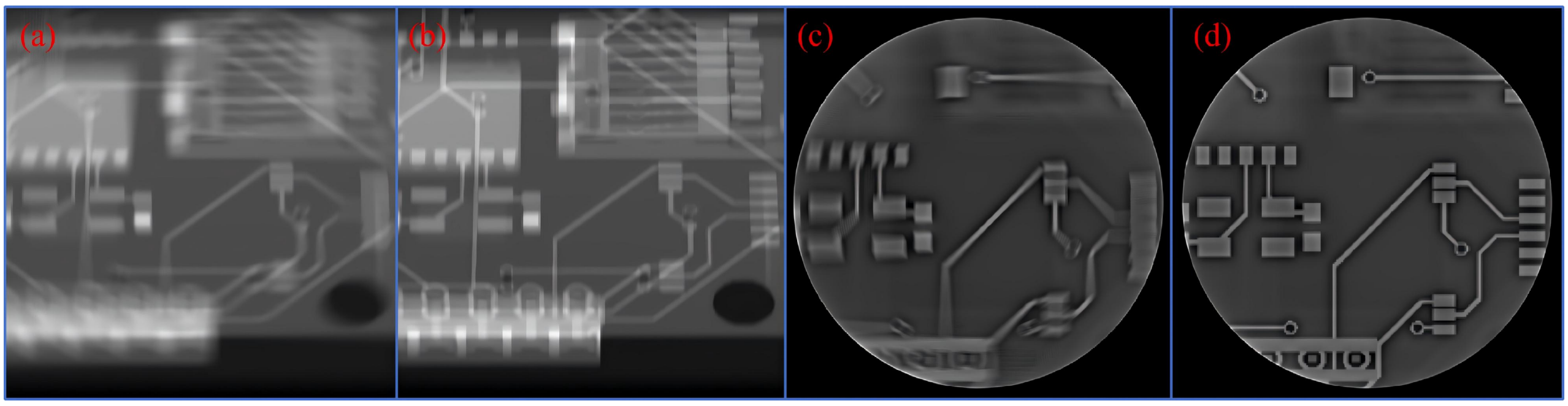

**Fig. 2.** Schematic diagram illustrating cross-domain rotational blur. (a) Blurry projection generated in the projection domain due to continuous exposure; (b) Clear projection acquired under ideal static scanning conditions; (c) Rotational blur in the reconstructed slice, introduced by the propagation of projection-domain blur during 3D reconstruction; (d) Clear reference ground-truth slice.

Furthermore, the rotational blur in RCL exhibits significant spatial variance, as illustrated in Fig. 3. For a specific point located at a radius r from the rotation center, the arc length of the blur trajectory formed within the exposure angular range $\Delta\theta$ can be approximated as:

$$L = r \cdot \Delta\theta = r \cdot (\omega \cdot \Delta T), \tag{8}$$

where $\omega$ is the angular velocity of rotation, and $\Delta T$ is the single exposure time. Equation (8) explicitly demonstrates that the length of the blur trajectory is positively correlated with the rotation radius, scanning angular velocity, and exposure time.

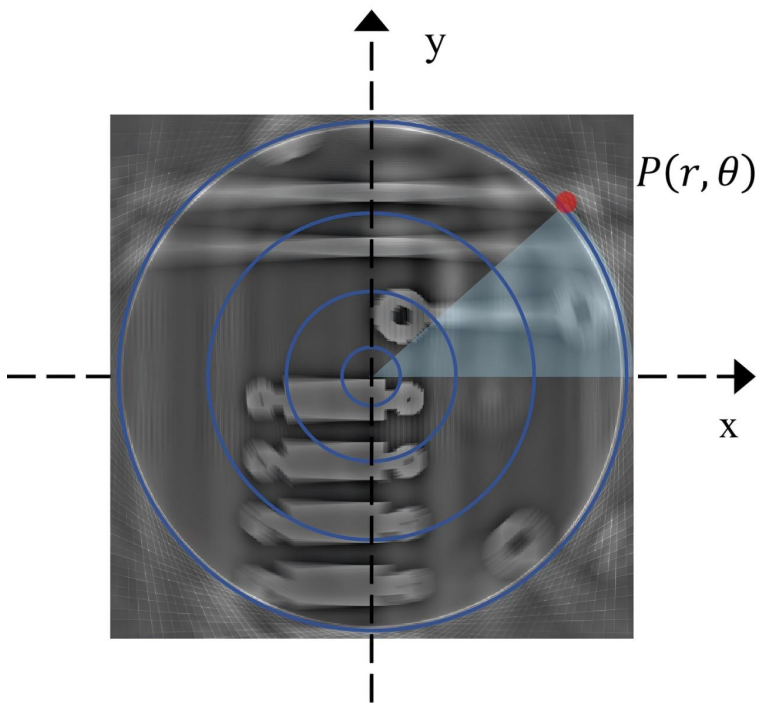

**Fig. 3.** Spatial variance of rotational blur in an RCL slice. The length of the arc-shaped blur trajectory increases linearly with the rotation radius r, indicating that peripheral regions far from the rotation center suffer from larger-scale and more severe non-uniform degradation.

Nevertheless, under sparse-views acquisition conditions, if the rotational blur is solely corrected in the projection domain to restore high-frequency features, the subsequent FDK reconstruction will inherently induce distinct sparse artifacts in the slice images due to severe data undersampling[35]. Let $\mathcal{R}^T_{sparse}$ and $\mathcal{R}^T_{continuous}$ denote the discrete reconstruction operator under actual sparse-view conditions and the ideal operator under continuous-view conditions, respectively. The induced sparse artifacts, denoted as $\mathcal{E}_{sparse}$ can be formulated as:

$$\mathcal{E}_{sparse} = \mathcal{R}^T_{sparse} P_{ideal} - \mathcal{R}^T_{continuous} P_{ideal}, \tag{9}$$

Here, $P_{ideal}$ represents the ideal clear projection data. When combined with $\mathcal{R}^T_{sparse}$, it refers to the ideal projections sampled at the same sparse angular positions as $P_{blur}$; when combined with $\mathcal{R}^T_{continuous}$, it refers to the complete ideal projection data under the continuous-view condition. Correspondingly, the operator $\mathcal{R}^T$ in Eq. (6) is equivalent to $\mathcal{R}^T_{continuous}$.

Under the same sparse-view reconstruction operator, the reconstruction-domain error propagated from projection-domain rotational blur is explicitly defined as

$$\mathcal{E}_{\text{rot}} = \mathcal{R}^T_{\text{sparse}} P_{\text{blur}} - \mathcal{R}^T_{\text{sparse}} P_{\text{ideal}} = \mathcal{R}^T_{\text{sparse}} \left(P_{\text{blur}} - P_{\text{ideal}}\right). \tag{10}$$

In summary, the degradation observed in the final reconstructed RCL slices does not originate from a singular factor; rather, it is a complex composite degradation comprising projection-domain rotational blur and image-domain sparse artifacts. Therefore, the actually degraded slice can be comprehensively expressed as:

$$I_{blur} = \mathcal{R}^T_{sparse} P_{blur} = \text{I}_{gt} + \mathcal{E}_{rot} + \mathcal{E}_{sparse}, \tag{11}$$

where $\mathcal{E}_{rot}$ represents the degradation term caused by the propagation of projection-domain rotational blur after reconstruction, and $\mathcal{E}_{sparse}$ represents the sparse artifacts induced by limited-view acquisition. As explicitly indicated by Equation (11), this degradation process exhibits distinct stage-wise and cross-domain coupling characteristics. This physical formulation provides a robust theoretical foundation for the targeted cross-domain decoupling and cascaded restoration strategy proposed in this study.

### 3.2 Overall Architecture

As illustrated in Fig. 4, the proposed RCL-Mamba employs a two-stage cascaded restoration architecture comprising a projection domain network (P-Net) and an image domain network (I-Net), which are interconnected via the FDK reconstruction operator. Both networks are constructed based on a hierarchical encoder-decoder framework.

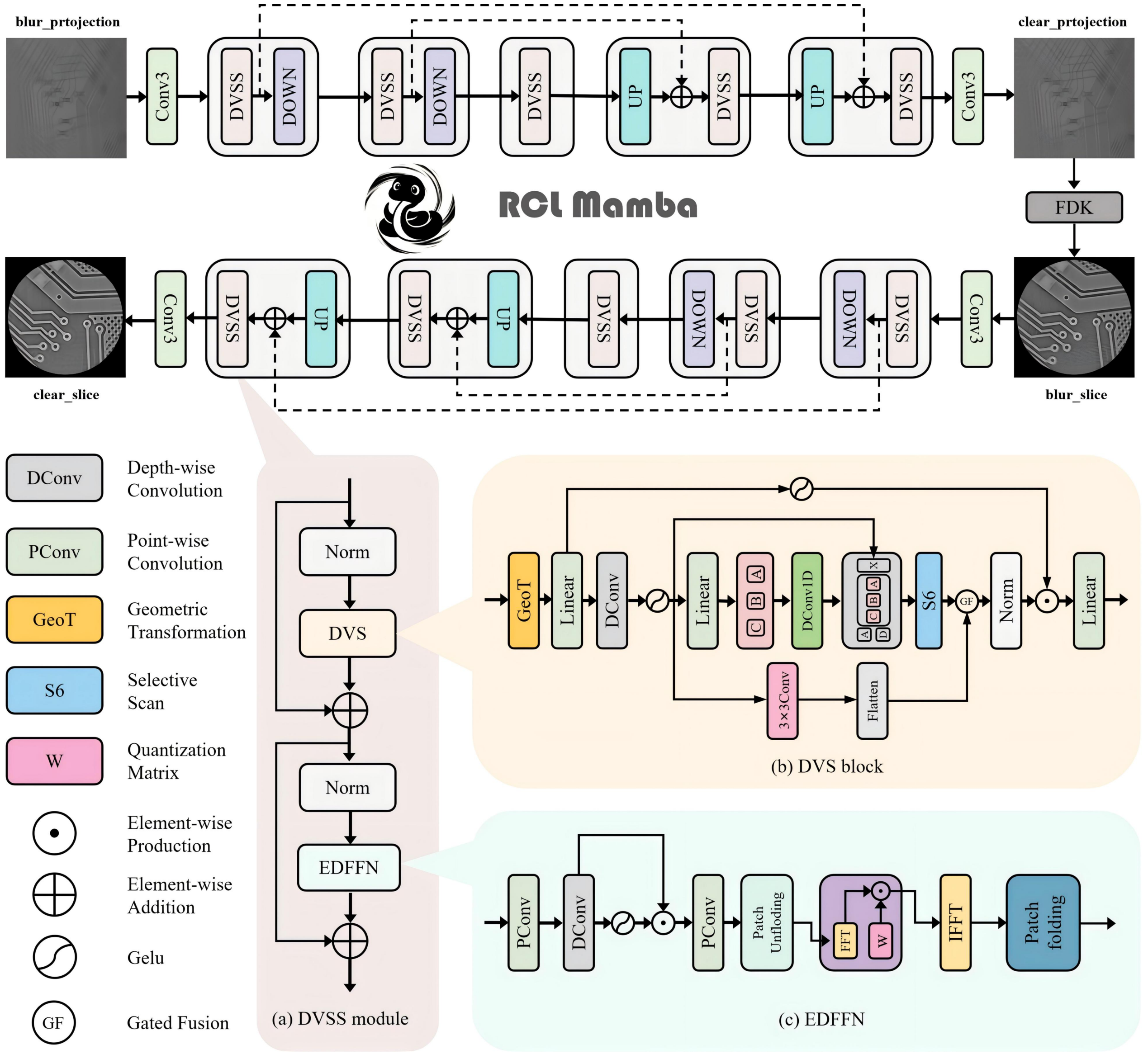

**Fig. 4.** The overall architecture of RCL-Mamba. (a) The Dual-branch Vision State Space (DVSS) module, serving as the fundamental feature extraction unit, primarily consists of a DVS block and an EDFFN [30] block; (b) The DVS block models global degradation trajectories and localized features through geometric transformations and a Mamba-CNN dual-branch structure; (c) The EDFFN [30] block incorporates frequency-domain transformations to further enhance the recovery of high-frequency microstructures.

Specifically, given a blurry projection $P_{blur} \in \mathbb{R}^{H\times W}$ , shallow features $F_s \in \mathbb{R}^{H\times W\times C}$ are first extracted through a $3\times 3$ convolution,where $H\times W$ denotes the spatial dimensions and $C$ represents the number of feature channels. Subsequently, $F_s$ enters a three-level symmetric encoder-decoder network. Each hierarchical level consists of multiple stacked Dual-branch Vision State Space (DVSS) modules. The feature dimensions vary progressively across the levels as $\mathbb{R}^{\frac{H}{2^{l-1}}\times\frac{W}{2^{l-1}}\times 2^{l-1}C}$ ,where $l=1, 2, 3$ . The network employs bilinear interpolation and $1\times 1$ convolutions to achieve upsampling and downsampling, respectively, and utilizes skip connections to preserve structural details. Upon traversing the hierarchical structure, P-Net outputs the restored projection $P_{restored}$ ,which is then reconstructed into a tomographic slice $I_{rec} \in \mathbb{R}^{H\times W}$ via the FDK algorithm. Finally, I-Net performs secondary refinement on $I_{rec}$ to suppress residual sparse artifacts, ultimately generating the high-fidelity clear slice $I_{final} \in \mathbb{R}^{H\times W}$ .

### 3.3 Projection-Image Dual-Domain Joint Processing

Due to the inherent cross-domain composite degradation characteristics of RCL, single-domain restoration paradigms frequently encounter optimization bottlenecks. To address this, a cascaded processing strategy across the projection and image domains is designed, effectively decomposing the complex restoration task into two targeted stages. First, P-Net, denoted as $\mathcal{N}_p$, is utilized to correct rotational blur. This targeted operation suppresses front-end degradation at the imaging source and fundamentally mitigates error propagation during the subsequent reconstruction. Subsequently, the restored projection $P_{restored}$ is reconstructed via FDK to obtain the initial slice $I_{rec}$. Finally, I-Net, denoted as $\mathcal{N}_I$, performs secondary refinement to efficiently suppress the residual sparse artifacts and local structural distortions within $I_{rec}$. This cascaded restoration process is mathematically formulated as follows:

$$P_{restored} = \mathcal{N}_p\left(P_{blur}\right), \tag{12}$$

$$I_{rec} = \mathcal{R}^T P_{restored}, \tag{13}$$

$$I_{final} = \mathcal{N}_I\left(I_{rec}\right), \tag{14}$$

Through this targeted cascaded decoupling, the proposed framework achieves progressive restoration and the synergistic suppression of cross-domain composite distortions.

### 3.4 Mamba-CNN Dual-Branch Module

To address the coexistence of large-scale rotational blur and local detail degradation in RCL, a Mamba-CNN dual-branch module is designed to synergize different inductive biases, facilitating the decoupling and fusion of global blur trajectories and local structural features.

The Mamba branch is responsible for global degradation modeling. To enhance the perception of rotational geometric directions, the input features $F_{in}$ first undergo a geometric transformation to obtain $G = GeoT\left(F_{in}\right)$. Here, $G = GeoT\left(\cdot\right)$ denotes an alternating geometric transformation: for the i-th DVSS module, the input feature is transposed when i is even and flipped along both the horizontal and vertical axes when i is odd. Both transformations are reversible, allowing the original spatial arrangement to be restored by the corresponding inverse transformation. Subsequently, discretization parameters are dynamically generated via a convolutional layer, and state-space recurrence is performed under the selective scan operator $S6$ to generate the global feature representation $F_{global} = S6(G, A, B, C, D, \Delta)$ [30] . Here, $A$ and $D$ are learnable parameter matrices. The dynamic parameters $B$, $C$, and the discretization step $\Delta$ are derived from the input via linear projections followed by a 1D depth-wise convolution . This design allows the model to incorporate spatial context when deriving these parameters, enabling them to better capture local and global dependencies before the selective scan.

Conversely, the CNN branch handles localized structural representation, extracting high-frequency textures and edge information. After geometric transformation, linear mapping, non-linear activation, and a $3\times3$ convolution, the local feature representation $F_{local}$, structurally aligned with the global branch, is generated:

$$F_{local} = Flatten\left(Conv_{3\times3}\left(GELU\left(DConv\left(Linear\left(GeoT\left(F_{in}\right)\right)\right)\right)\right)\right) \tag{15}$$

To mitigate information conflict caused by simple feature concatenation, the module introduces a dynamic gated feature fusion mechanism. After concatenating $F_{global}$ and $F_{local}$ in the channel dimension, a spatially adaptive gating weight M is generated through linear mapping and a Sigmoid activation function $\sigma$ :

$$M = \sigma\left(Linear\left(\left[F_{global}, F_{local}\right]\right)\right), \tag{16}$$

based on this weight, features from both branches are adaptively fused and processed through a normalization layer to stabilize the feature distribution, yielding the final output feature $F_{out}$ :

$$F_{out} = Norm\left(M \odot F_{local} + \left(1 - M\right) \odot F_{global}\right), \tag{17}$$

where $\odot$ denotes element-wise multiplication. This mechanism adaptively adjusts the contribution ratio of the global and local branches based on the degree of degradation at different spatial locations, achieving adaptive coordination between large-scale blur correction and local detail preservation.

### 3.5 Loss Function

To describe the optimization objectives of the two sequential training stages, the same composite loss consisting of pixel-domain, frequency-domain, and edge losses is employed in both stages. Let $X_{\text{out}}$ and $X_{gt}$ denote the restored output and its corresponding ground truth in the current stage, respectively. For the projection-domain stage, $X_{\text{out}} = P_{restored}$ and $X_{\text{gt}} = P_{gt}$ ; for the image-domain stage, $X_{\text{out}} = I_{final}$ and $X_{gt} = I_{gt}$

First, the $L_1$ loss is employed to constrain the discrepancy between the restored output and the ground truth in the pixel domain. The pixel-domain loss $\mathcal{L}_{pix}$ is formulated as：

$$\mathcal{L}_{pix} = \left\| X_{out} - X_{gt} \right\|_1 \tag{18}$$

Second, a frequency-domain loss $\mathcal{L}_{fft}$ is introduced to enhance the model's capability to recover fine-grained textures and high-frequency microstructures:

$$\mathcal{L}_{fft} = \left\| \mathcal{F}\left(X_{\text{out}}\right) - \mathcal{F}\left(X_{gt}\right) \right\|_1 , \tag{19}$$

where $\mathcal{F}(\cdot)$ denotes the discrete Fourier transform.

Furthermore, to suppress the over-smoothing phenomenon during the restoration process, an edge loss $\mathcal{L}_{edge}$ is introduced to constrain the structural edge responses:

$$\mathcal{L}_{edge} = \left\| \nabla\left(X_{out}\right) - \nabla\left(X_{gt}\right) \right\|_1 , \tag{20}$$

where $\nabla(\cdot)$ represents the edge extraction operator.

Ultimately, the overall loss function $\mathcal{L}$ for each stage is defined as:

$$\mathcal{L} = \mathcal{L}_{pix} + \alpha \mathcal{L}_{fft} + \beta \mathcal{L}_{edge} , \tag{21}$$

where $\alpha$ and $\beta$ are the weight coefficients for the frequency-domain loss and the edge loss, respectively, both of which are empirically set to 0.1 in this paper.

## 4. Experiments and Results

### 4.1 Dataset

#### 4.1.1 Simulation Dataset

To address the scarcity of high-quality paired real-world data, a tomographic imaging simulation framework was constructed. 28 industrial-grade PCB 3D digital phantoms were utilized, with each model uniformly partitioned into four independent sub-regions in space, as illustrated in Fig. 5.

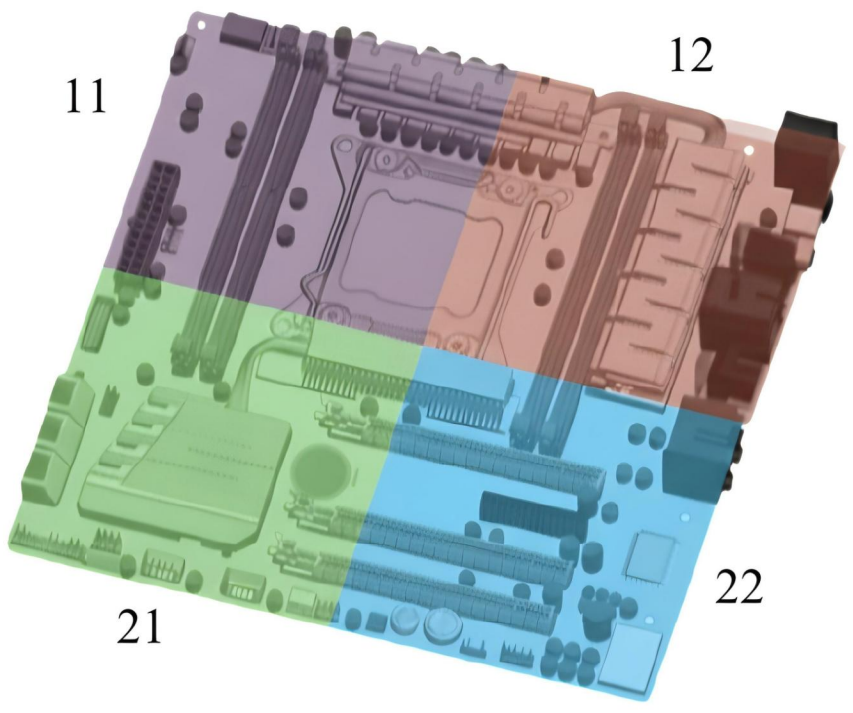

**Fig. 5.** Schematic of industrial-grade PCB 3D digital phantoms and sub-region partitioning. The entire PCB model is uniformly partitioned into four independent sub-regions: 11, 12, 21, and 22 in the spatial dimension.

Subsequently, projection simulations were performed using the TIGRE toolbox[36], with the geometric configuration detailed in Table 1. In this configuration, DSD denotes the distance from the source to the detector, DSO represents the distance from the source to the object, nDetector refers to the detector pixels, and θ signifies the laminography tilt angle. Additionally, view denotes the number of sampling views, sparse-view indicates the sparse sampling views, N represents the downsampling factor, and M refers to the number of left and right projection frames adjacent to the central view.

**Table. 1. Parameters for simulation dataset preparation**

| Parameter | DSD | DSO | nDetector | θ | view | sparse_view | N | M |
|---|---|---|---|---|---|---|---|---|
| Value | 600mm | 150mm | 512×512 | 45° | 512 | 64 | 8 | 4 |

As shown in Fig. 6, 512 instantaneous clear projections were first generated for each scanning region. To simulate the sparse scanning and exposure integration effects inherent in practical high-throughput inspection, 64-view blurry observation data and the corresponding ground truth (GT) were synthesized. Crucially, this 64-view sparse sampling configuration mathematically compresses the physical scanning time from 102.4 s to 12.8 s.

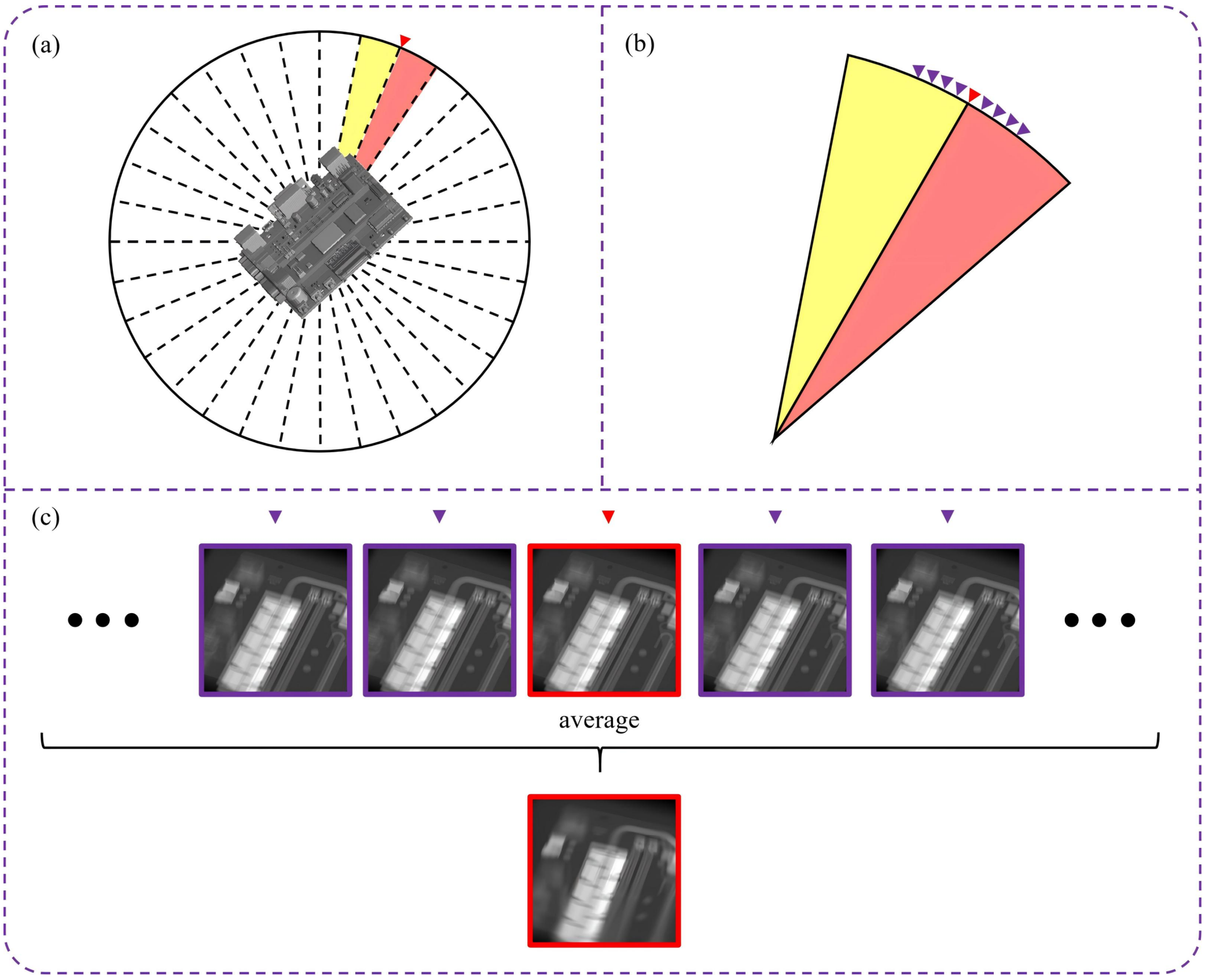

**Fig. 6.** Illustration of blurry projection generation under the RCL dynamic sparse-view scanning mode. With the downsampling factor set to N=8,64 target views are uniformly extracted from 512 full views. Centered at each target view, its adjacent M=4 instantaneous projections on both sides (9 frames in total) are symmetrically fused to generate the final 64-view blurry observation data. The corresponding ground truth (GT) consists of 64 clear projections at the same sampling positions.

The dataset was systematically partitioned according to the strict requirements for dual-domain synergistic training. In the projection-domain stage, 10 phantoms were selected to construct the training set (totaling 2,560 paired samples: 10 phantoms × 4 regions × 64 views ). In the image-domain stage, slices were reconstructed from the restored projections via the FDK algorithm, and key layers from 16 phantoms were extracted to form the training set (totaling 2,560 paired slices). The test set comprises two independent phantoms distinct from the training data, yielding 512 pairs of projection test samples and 320 pairs of slice test samples. To ensure a fair comparison, all single-domain baseline methods were strictly trained using the identical set of 4,160 image-domain slices and evaluated on the exact same 320 slice pairs.

### 4.1.2 Real-world Dataset

To rigorously evaluate the generalization capability of the proposed model in practical industrial environments, a real-world dataset was constructed utilizing a physical RCL inspection system. This system employed a 64-view rapid scanning mode to acquire the degraded blurry observation data, while the corresponding high-fidelity ground truth was established using reconstruction results derived from a 512-view dense scan.The photograph of the experimental setup is shown in Fig. 7.The partitioning logic for the real-world dataset was maintained consistent with that of the simulation experiments. Specifically, in the projection-domain stage, 5 physical PCB phantoms were selected to construct the training set (totaling 1,280 paired projections). In the image-domain stage, 8 phantoms were utilized to form the training set (totaling 1,280 paired slices). The test set comprised 2 independent phantoms strictly distinct from the training data, yielding 512 pairs of projection test samples and 320 pairs of slice test samples. Furthermore, to ensure fairness in comparative evaluations, the training and testing phases for all single-domain baseline methods strictly adhered to this identical image-domain slice dataset configuration, with no additional auxiliary data introduced.

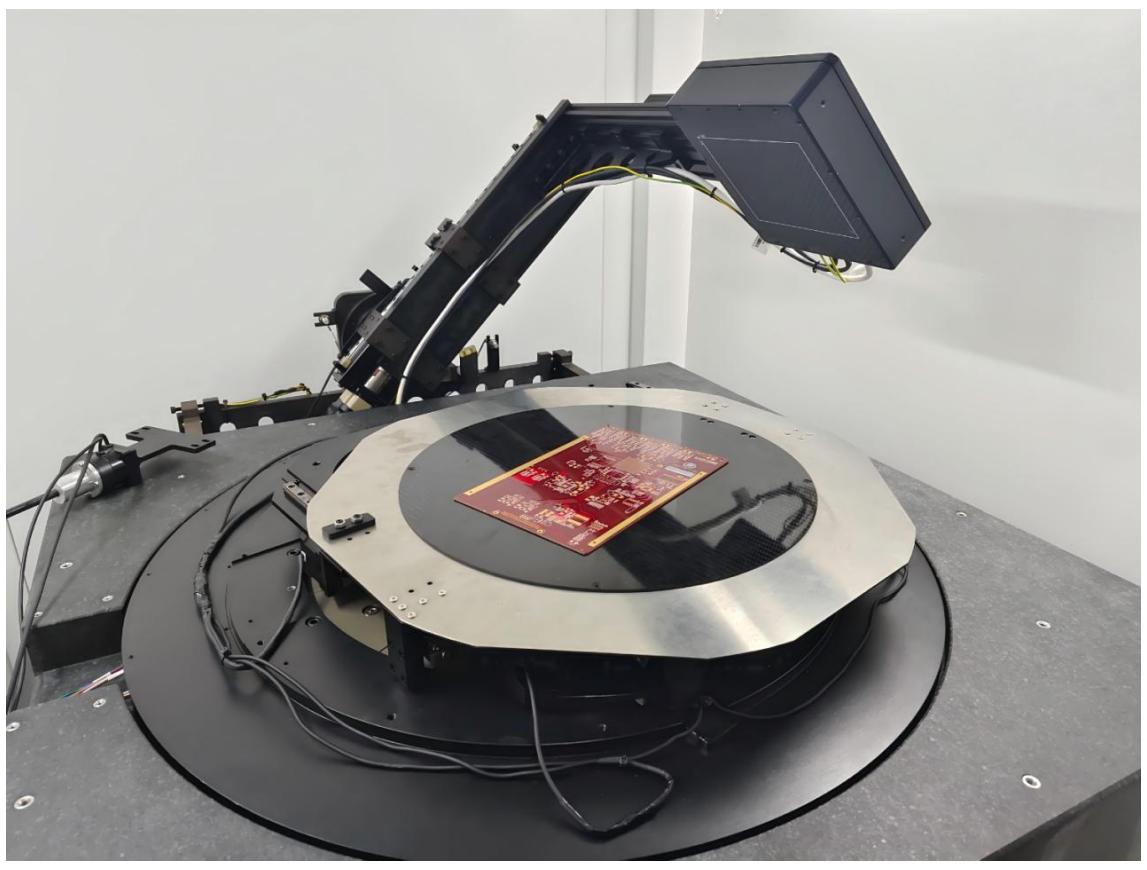

**Fig. 7.** Photograph of the physical RCL inspection system utilized for real-world data acquisition.

### 4.1.3 Implementation Details

A stage-wise training strategy was adopted to optimize P-Net and I-Net sequentially. Specifically, P-Net was first trained using paired blurry and clear projections. After convergence, the parameters of P-Net were fixed, and the restored projections were reconstructed using the FDK algorithm to generate the training inputs for I-Net. I-Net was then trained separately, without alternating or end-to-end joint optimization. Both stages utilized the same network backbone, loss function, and AdamW optimizer. Regarding the network configuration, the number of channels in the shallow feature extraction layer was set to 48, and the DVSS modules were stacked in the three encoder-decoder levels with a distribution of [6, 6, 12]. To enhance training stability, a progressive training strategy was employed: the first stage involved 300,000 iterations on $128\times128$ pixel image patches with a batch size of 8, where the learning rate decayed from $1\times10^{-3}$ to $1\times10^{-7}$ following a cosine annealing strategy. In the second stage, the patch size was increased to $256\times256$ pixels with a batch size of 2, and the training continued for another 300,000 iterations with the learning rate decaying from $5\times10^{-4}$ to $1\times10^{-7}$. Random flipping and rotation were introduced during training for data augmentation. The total training time was approximately 176 hours. Peak Signal-to-Noise Ratio (PSNR), Structural Similarity Index Measure (SSIM)[37], and Gradient Magnitude Similarity Deviation (GMSD)[38] were adopted as the evaluation metrics. Note that a lower GMSD value indicates higher perceptual image quality and better preservation of local structural gradients. The comparison baselines included the traditional FDK[34] method, CNN-based methods, i.e., NAFNet [14] and MLWNet-B [15], Transformer-based methods, i.e., Restormer[18], FFTformer[21], and MDT[22], and Mamba-based methods, i.e., EAMamba[29] and MaIR[32]. All comparison methods were reproduced using their official open-source implementations and the training configurations recommended in the corresponding papers. The method-specific loss functions, learning-rate schedules, and training iterations were retained to faithfully reproduce the original methods. Each model was trained until its validation performance converged, and the checkpoint achieving the best validation performance was selected for final testing. The proposed model was implemented using PyTorch, and all training and testing experiments were conducted on a single NVIDIA RTX 4090 GPU with 24 GB of memory.The detailed hyperparameter configurations for each method are summarized in Table 2.

**Table. 2.** Summary of key parameter settings for the comparison methods.

| Method | Dims | Blocks | Refine. | Learning Rate | Scheduler Strategy | Iters | Loss Function |
|---|---|---|---|---|---|---|---|
| NAFNet[14] | 64 | Enc:[1, 1, 1, 28]<br>Dec:[1, 1, 1, 1] | - | $1\times10^{-3}$ | Cosine Annealing | 400k | PSNR |
| MDT[22] | 48 | [6, 6, 12] | 4 | $1\times10^{-3}$ | Cosine Annealing | 300k | $L_1$ + FFT |
| MLWNet-B[15] | 64 | - | - | $9\times10^{-4}$ | Cosine Annealing | 800k | SRN +Wavelet |
| Restormer[18] | 48 | [4, 6, 6, 8] | 4 | $3\times10^{-4}$ | Restart Cyclic | 300k | $L_1$ |
| FFTformer[21] | 48 | [6, 6, 12] | 4 | $1\times10^{-3}$ | Cosine Annealing | 300k | $L_1$ + FFT |
| EAMamba[29] | 64 | [4, 6, 6, 7] | 2 | $3\times10^{-4}$ | Restart Cyclic | 450k | - |
| MaIR[32] | 48 | [4, 6, 6, 8] | 4 | $3\times10^{-4}$ | Restart Cyclic | 300k | $L_1$ |

### 4.2 Simulation Experiment

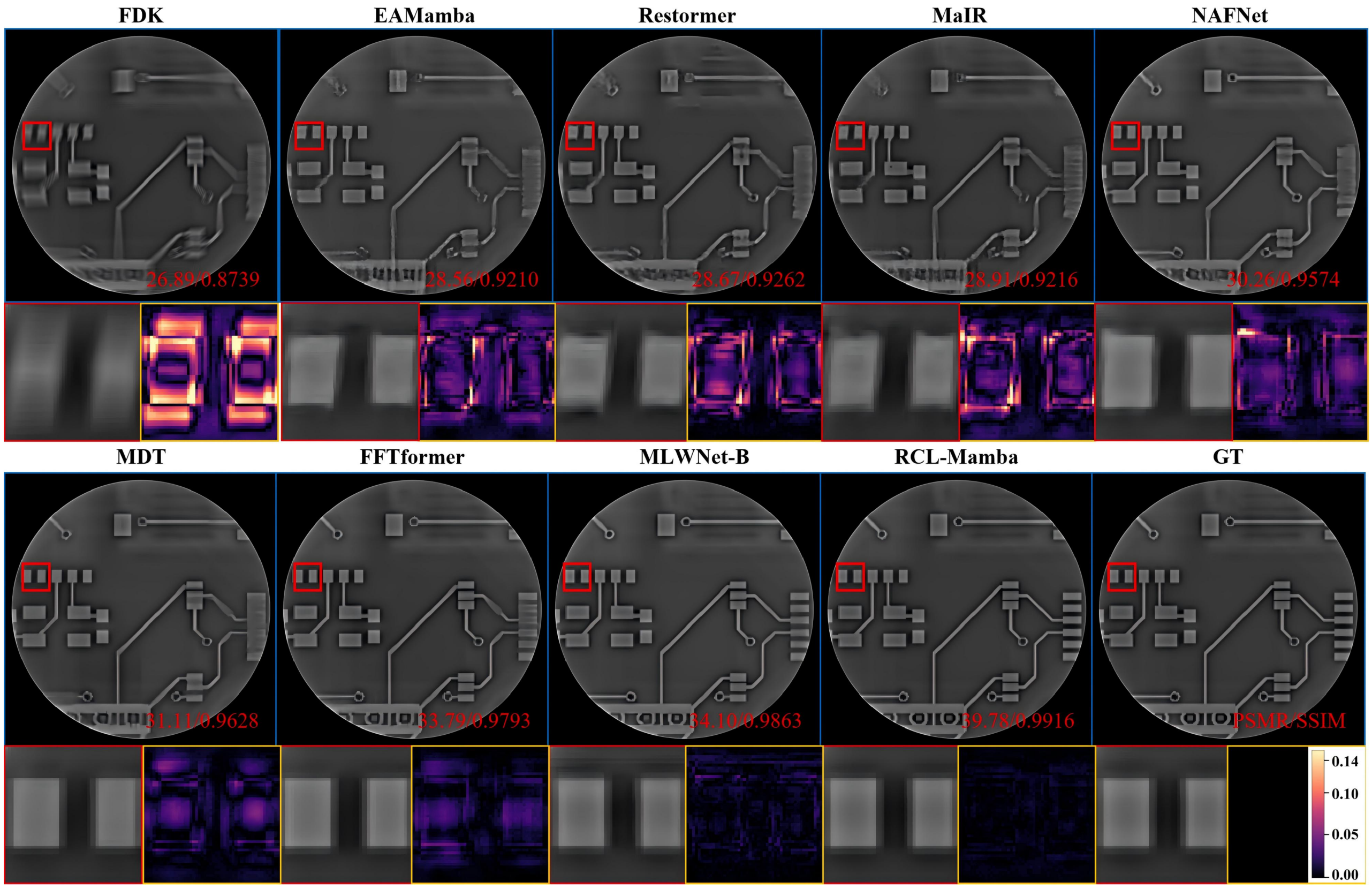

**Fig. 8.** Qualitative reconstruction results on the simulated dataset. Top row from left to right: FDK, EAMamba, Restormer, MaIR, and NAFNet. Bottom row from left to right: MDT, FFTformer, MLWNet-B, RCL-Mamba (ours), and ground truth. The first row shows the full reconstructions with regions of interest (ROIs) indicated by red solid boxes. The second row provides the corresponding ROIs (framed in red) alongside the error residual heatmaps (framed in yellow). All reconstructed images are displayed using a normalized intensity window of [0, 1].

Fig. 8 provides a qualitative comparison of the reconstructed results. The analytical FDK [34] reconstruction exhibits severe rotational blur and sparse artifacts, lacking prior regularization for highly incomplete data. Single-domain global models also struggle. The Transformer-based Restormer [18] produces severe ghosting because operating solely in the image domain fails to decouple the cross-domain composite degradation. Similarly, single-domain Mamba models (EAMamba [29] and MaIR [32]) leave uncorrected arc-shaped blur. This failure stems from their fixed linear scan patterns over 2D feature maps, which rigidly disrupt the strongly anisotropic, arc-shaped topology inherent in RCL trajectories. Other baselines show progressive but limited results. NAFNet [14] leaves noticeable residual blur because its constrained local receptive fields cannot capture large-scale, spatially-variant degradation. MDT [22] and FFTformer [21] achieve more coherent global restoration but over-smooth pad boundaries, as their single-domain nature forces a compromise between artifact suppression and high-frequency edge preservation. The CNN-based MLWNet-B [15] achieves a highly competitive sub-optimal result, removing most rotational blur thanks to multi-level wavelet transforms bolstering context modeling. However, minor edge smoothing persists due to the spatial stationarity assumption of local convolutions. In striking contrast, the proposed RCL-Mamba achieves the highest visual fidelity, sharply recovering the rectangular pads without visible ghosting. Its nearly black error heatmap confirms the minimum prediction deviation. This performance advantage is driven by the dual-domain cascaded strategy, which prevents cross-domain error propagation by directly addressing degradation in the physical sequence. Furthermore, unlike the fixed linear scan patterns over 2D feature maps adopted by existing Mamba baselines, our Mamba branch explicitly introduces feature map flipping and inversion operations. This critical design aligns the state-space recurrence with the arc-shaped geometric topology to effectively eliminate ghosting, while the parallel CNN branch concurrently ensures high-frequency local detail recovery.

Table 3 presents the quantitative comparison results. RCL-Mamba achieves the best performance across all three metrics, with a PSNR of 38.46 dB, an SSIM of 0.9862, and a GMSD of 0.0099. Compared to the second-best method (MLWNet-B), RCL-Mamba yields a substantial PSNR gain of 2.98 dB. The exceptionally low GMSD score further indicates that our method effectively preserves the gradient magnitude of local structures, consistent with the visual clarity observed in Fig. 8.

**Table. 3. Quantitative evaluation of the simulated rotational blur dataset in terms of PSNR, SSIM, and GMSD**

| Methods | PSNR(dB) | SSIM | GMSD |
|---|---|---|---|
| FDK[34] | 28.57 | 0.9087 | 0.0850 |
| NAFNet[14] | 34.91 | 0.9753 | 0.0358 |
| MDT[22] | 33.62 | 0.9728 | 0.0241 |
| MLWNet-B[15] | 35.48 | 0.9811 | 0.0146 |
| Restormer[18] | 32.27 | 0.9586 | 0.0605 |
| FFTformer[21] | 34.26 | 0.9766 | 0.0177 |
| EAMamba[29] | 31.02 | 0.9514 | 0.0737 |
| MaIR[32] | 32.96 | 0.9618 | 0.0531 |
| RCL-Mamba | 38.46 | 0.9862 | 0.0099 |

### 4.3 Real-world Experiments

Fig. 9 provides qualitative comparisons on the real-world dataset. The FDK [34] reconstruction completely distorts the circular via, inherently lacking regularization to suppress uncalibrated measurement noise and sparse-view artifacts. Restormer [18], as a self-attention-based global restoration model, suffers from evident geometric distortion under real-world interference, indicating that global token aggregation can easily propagate distant noisy responses and compromise local structural consistency. The Mamba-based methods, EAMamba [29] and MaIR [32], also fail to preserve the circular topology, degenerating the via into an irregular blob. This suggests that their fixed linear scan patterns over 2D feature maps tend to incorrectly entangle spatially distant noisy pixels, thereby destroying fine local geometries. NAFNet [14] leaves traces heavily blurred, as its restricted receptive fields cannot resolve large-scale rotational blur. MDT [22] and FFTformer [21] mitigate global artifacts but distort circular boundaries, forcing a sub-optimal trade-off between noise suppression and high-frequency geometric preservation. The CNN-based MLWNet-B [15] achieves the sub-optimal result by reconstructing a recognizable via, benefiting from multi-level wavelets enhancing context modeling. However, high-frequency edge loss persists, as local convolutions inevitably over-smooth sharp boundaries when combatting severe measurement noise. In contrast, RCL-Mamba faithfully reconstructs the precise circular topology, confirmed by its nearly black error heatmap. This superiority is jointly driven by two factors: the dual-domain cascaded strategy directly filters physical measurement interference in the projection domain before it propagates into image-domain artifacts, while the Mamba-CNN dual-branch module optimally preserves local micro-structure geometries against severe real-world degradation.

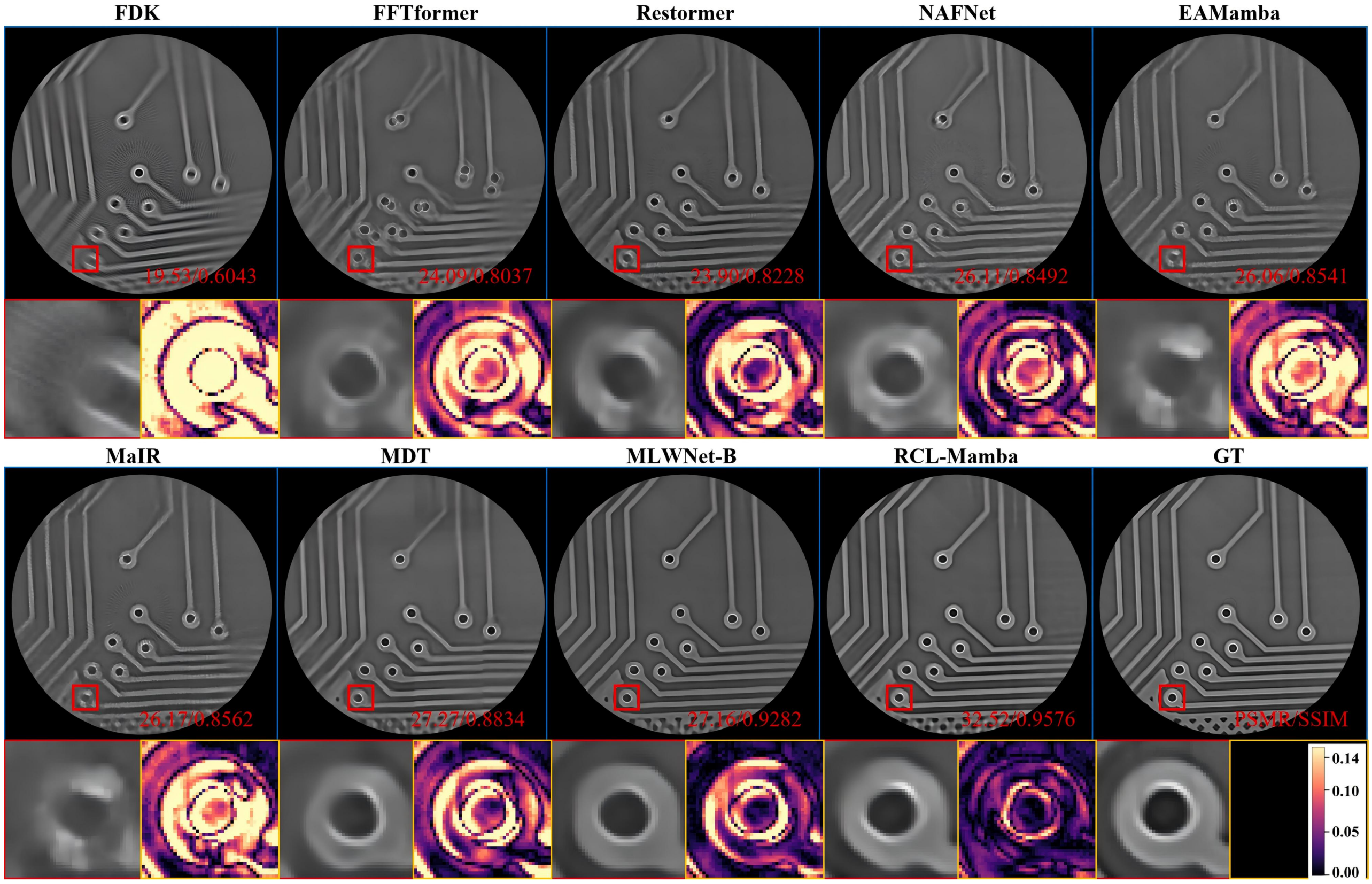

**Fig. 9.** Qualitative reconstruction results on the real-world dataset. Top row from left to right: FDK, FFTformer, Restormer, NAFNet, and EAMamba. Bottom row from left to right: MaIR, MDT, MLWNet-B, RCL-Mamba (ours), and ground truth. The first row shows the full reconstructions with regions of interest (ROIs) indicated by red solid boxes. The second row provides the corresponding ROIs (framed in red) alongside the error residual heatmaps (framed in yellow). All reconstructed images are displayed using a normalized intensity window of [0, 1].

Table 4 presents the quantitative results. RCL-Mamba achieves the best performance with a PSNR of 32.47 dB, an SSIM of 0.9543, and a GMSD of 0.0464, outperforming the second-best method (MLWNet-B) by 3.80 dB in PSNR. The lowest GMSD further confirms the superior preservation of local structural gradients, consistent with the visual fidelity observed in Fig. 9.

**Table. 4.** Quantitative evaluation of the real-world rotational blur dataset in terms of PSNR, SSIM, and GMSD

| Methods | PSNR(dB) | SSIM | GMSD |
|---|---|---|---|
| FDK[34] | 22.05 | 0.7511 | 0.1379 |
| NAFNet[14] | 27.14 | 0.9060 | 0.0823 |
| MDT[22] | 26.52 | 0.9047 | 0.0856 |
| MLWNet-B[15] | 28.67 | 0.9362 | 0.0667 |
| Restormer[18] | 26.35 | 0.8950 | 0.0958 |
| FFTformer[21] | 25.05 | 0.8664 | 0.1137 |
| EAMamba[29] | 27.00 | 0.8938 | 0.0880 |
| MaIR[32] | 27.13 | 0.9027 | 0.0869 |
| RCL-Mamba | 32.47 | 0.9543 | 0.0464 |

## 4.4 Ablation Study

To verify the effectiveness of the proposed core modules, ablation experiments were conducted on both the simulated and real-world datasets, focusing on the individual contributions of the projection－image dual-domain cascade strategy and the Mamba-CNN dual-branch module. All ablation models were evaluated using the identical training configuration as the primary experiments.

Table 5 presents the quantitative comparison results of the various ablation configurations. In this context, "Baseline" refers to the network performing single-stage restoration solely in the image domain; "Double-Stage" indicates the introduction of the cascade processing strategy; "Mamba-CNN" represents the inclusion of the dual-branch feature modeling module; and "RCL-Mamba" denotes the complete model incorporating both innovative designs.

First, the effectiveness of dual-domain joint processing is validated by comparing the Baseline and Double-Stage models. As illustrated in Fig. 10, upon introducing the cascade strategy, the residual sparse artifacts are effectively eliminated, and the overall visual fidelity and structural consistency of the slices are significantly enhanced. Quantitatively, as shown in Table 5, the dual-domain cascade strategy leads to a substantial PSNR improvement of 1.58 dB (reaching 38.04 dB) and a marked reduction in GMSD from 0.0125 to 0.0106. This result demonstrates that correcting rotational blur in the front-end projection domain fundamentally reduces the accumulation and propagation of sparse artifacts induced during reconstruction.

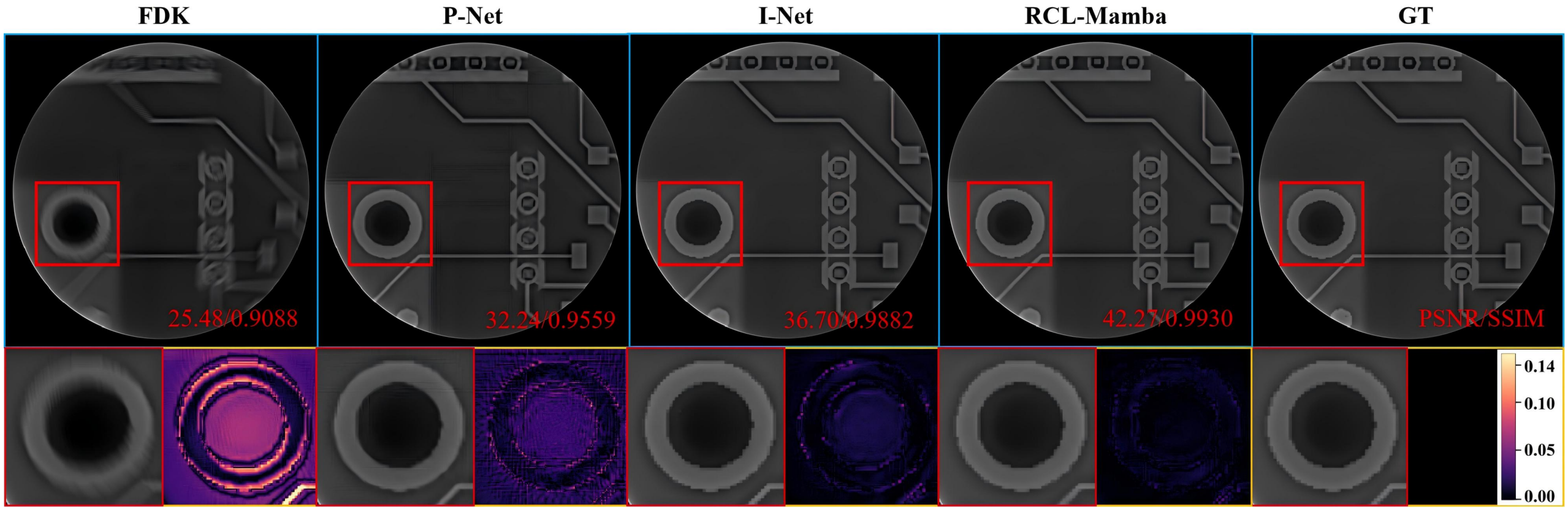

**Fig. 10.** Visual ablation results for the dual-domain cascade strategy. From left to right: FDK, P-Net (FDK reconstruction from the projections restored by P-Net), I-Net (image-domain-only baseline), RCL-Mamba (dual-domain), and ground truth. The first row shows the full reconstructions with regions of interest (ROIs) indicated by red solid boxes. The second row provides the corresponding ROIs (framed in red) alongside the error residual heatmaps (framed in yellow). All reconstructed images are displayed using a normalized intensity window of [0, 1].

Second, the contribution of the Mamba-CNN dual-branch module is examined. As evidenced by the local magnifications and residual heatmaps in Fig. 11, the complete RCL-Mamba demonstrates visibly clearer high-frequency edge restoration at the boundaries of slender traces and vias, indicating that the dual-branch architecture effectively synergizes CNN-based localized representation with Mamba-based global degradation modeling. Quantitatively, as shown in Table 5, applying this module to the Baseline improves PSNR by 0.31 dB (to 36.77 dB); when integrated with the dual-domain strategy, it yields an additional 0.42 dB gain over the Double-Stage model, reaching 38.46 dB with the optimal GMSD of 0.0099. The consistent improvement across both configurations confirms the effectiveness of the dynamic gated fusion mechanism in balancing global blur correction with local detail preservation.

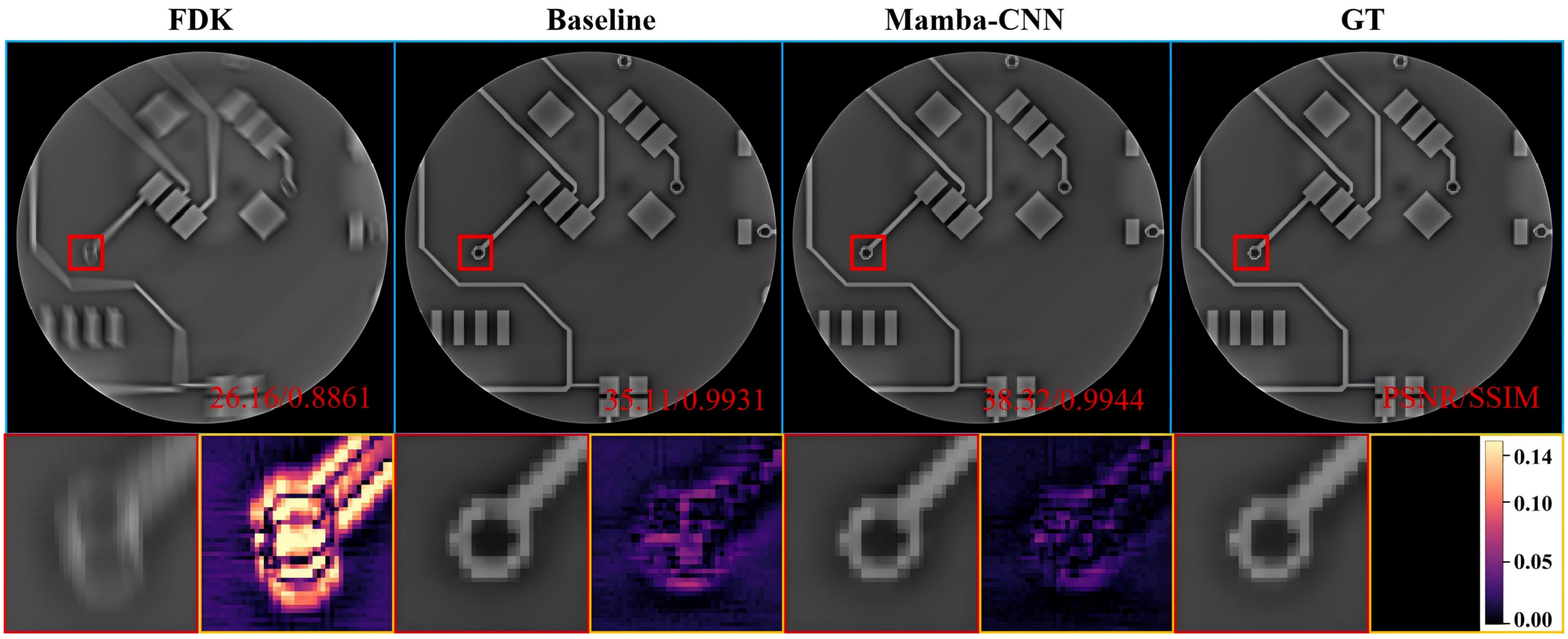

**Fig. 11.** Visual ablation results for the Mamba-CNN dual-branch module. The first row shows the full reconstructions with regions of interest (ROIs) indicated by red solid boxes. The second row provides the corresponding ROIs (framed in red) alongside the error residual heatmaps (framed in yellow). All reconstructed images are displayed using a normalized intensity window of [0, 1].

**Table. 5.** Quantitative evaluation of ablation experiments on the simulated dataset in terms of PSNR, SSIM, and GMSD

| Methods | Double-Stage | Mamba-CNN | PSNR | SSIM | GMSD |
|---|---|---|---|---|---|
| Baseline | | | 36.46 | 0.9810 | 0.0125 |
| Double-Stage | √ | | 38.04 | 0.9857 | 0.0106 |
| Mamba-CNN | | √ | 36.77 | 0.9826 | 0.0121 |
| RCL-Mamba | √ | √ | 38.46 | 0.9862 | 0.0099 |

To further verify the effectiveness of the proposed components under practical imaging conditions, additional ablation experiments were conducted on the real-world dataset.Overall, the quantitative results demonstrate that the two proposed components contribute to performance improvement at different levels. As shown in Table 6, compared with the Baseline, the Double-Stage strategy increases the PSNR from 28.74 to 32.11 dB, improves the SSIM from 0.9305 to 0.9536, and reduces the GMSD from 0.0752 to 0.0480, indicating that the projection - image domain cascade is the primary contributor to the suppression of rotational blur and reconstruction artifacts. Independently introducing the Mamba-CNN module also improves the PSNR by 0.39 dB and reduces the GMSD from 0.0752 to 0.0726, confirming its effectiveness in enhancing feature representation and preserving local structural details. When the Mamba-CNN module is further incorporated into the Double-Stage framework, the resulting RCL-Mamba achieves an additional PSNR gain of 0.36 dB, together with further improvements in SSIM and GMSD. Consequently, the complete model improves the PSNR by 3.73 dB over the Baseline and achieves the best overall performance of 32.47 dB, 0.9543, and 0.0464 in terms of PSNR, SSIM, and GMSD, respectively. These results verify that the dual-domain cascade strategy and the Mamba-CNN module provide complementary benefits, with the former mainly correcting global degradation and the latter further refining structural details.

**Table. 6.** Quantitative ablation results on the real-world dataset in terms of PSNR, SSIM, and GMSD.

| Methods | Double-Stage | Mamba-CNN | PSNR | SSIM | GMSD |
|---|---|---|---|---|---|
| Baseline | | | 28.74 | 0.9305 | 0.0752 |
| Double-Stage | √ | | 32.11 | 0.9536 | 0.0480 |
| Mamba-CNN | | √ | 29.13 | 0.9328 | 0.0726 |
| RCL-Mamba | √ | √ | 32.47 | 0.9543 | 0.0464 |

## 5. Discussion

### 5.1 Advantages of RCL-Mamba

The advantages of the proposed RCL-Mamba framework are primarily manifested across three dimensions: physical mechanism alignment, adaptive degradation modeling, and engineering efficiency. First, the dual-domain cascaded architecture facilitates the step-by-step decoupling of cross-domain composite degradations. By executing targeted restoration in both the projection and image domains, the framework avoids the ill-posed optimization bottlenecks inherent in single-domain paradigms. This approach aligns more rigorously with the underlying physical logic of RCL imaging, fundamentally mitigating error propagation during reconstruction. Second, the Mamba-CNN dual-branch module achieves the synergistic recovery of spatially-variant, long-range blur trajectories and localized high-frequency textures while strictly maintaining linear computational complexity. This dual-path modeling strategy demonstrates exceptional robustness in preserving the structural fidelity of micro-features, such as PCB metal vias and slender traces, which are critical for defect identification. Finally, RCL-Mamba significantly optimizes engineering efficiency. While maintaining high-quality reconstruction results, the model drastically compresses the number of required scanning views. This reduction effectively shortens the physical scanning time from 102.4 s to 12.8 s per station, successfully satisfying the stringent dual requirements of high-throughput industrial NDT for both superior imaging quality and rapid inspection throughput.

### 5.2 Radial-Position-Dependent Restoration Performance

As indicated by Eq. (8), the length of the rotational blur trajectory increases with the distance from the rotation center. Therefore, global image-quality metrics alone cannot fully characterize the spatially variant restoration performance. To further evaluate the robustness of different methods at different radial positions, the valid circular field of view was divided into four concentric annular regions according to the normalized radius $\rho = r/R$, where $r$ denotes the distance from a pixel to the rotation center and $R$ denotes the radius of the valid field of view. As illustrated in Fig. 12(a), the four radial regions were defined as $R_1:[0,0.25)$, $R_2:[0.25,0.5)$, $R_3:[0.50,0.75)$, and $R_4:[0.75,1.00)$. PSNR and SSIM were then calculated independently within each annular region.

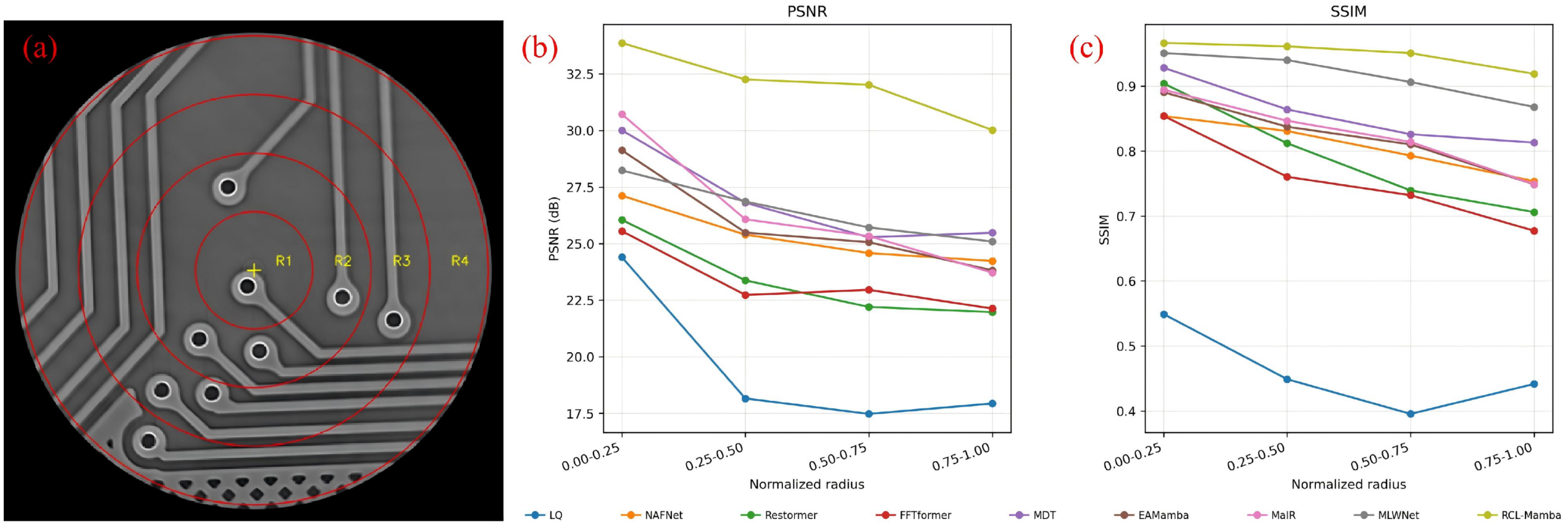

**Fig. 12.** Radial-position-dependent restoration performance on a representative real-world PCB slice. (a) Division of the valid circular field of view into four concentric annular regions according to the normalized radius; (b) PSNR comparison across different radial regions; (c) SSIM comparison across different radial regions. The reconstructed slice in (a) is displayed using a normalized intensity window of [0, 1].

As shown in Fig. 12(b) and Fig. 12(c), the restoration performance of most methods generally decreases as the normalized radius increases, which is consistent with the spatially variant rotational blur described in Eq. (8). The degradation is more pronounced in the peripheral regions because pixels farther from the rotation center undergo longer arc-shaped displacements during continuous exposure. In contrast, RCL-Mamba consistently achieves the highest PSNR and SSIM in all four radial regions. Its PSNR decreases moderately from 33.87 dB in $R_1$ to 30.02 dB in $R_4$, while its SSIM remains above 0.91 throughout the evaluated field of view. In the outermost region $R_4$ , RCL-Mamba exceeds the second-best method by 4.54 dB in PSNR and 0.0510 in SSIM. These results demonstrate that the proposed dual-domain cascaded restoration strategy remains effective in peripheral regions affected by more severe rotational blur and provides improved spatial robustness across the entire field of view.

### 5.3 Line-profile-based Structural Measurement

To further assess the local structural measurement fidelity of the reconstructed real-world PCB slices, a line-profile analysis was conducted, as shown in Fig. 13. Two representative profile lines were selected in the reconstructed slice. Specifically, L1 crosses a typical via/pad region and is used to evaluate the recovery of circular boundaries and local peak–valley responses, while L2 crosses a slender trace region and is used to evaluate trace-edge localization and gray-level transition consistency. For each profile, gray values were sampled along the selected line and averaged within a 5-pixel-wide band perpendicular to the profile direction, thereby reducing the influence of local noise.

As shown in Fig. 13(b), in the via/pad region, the LQ profile exhibits evident peak–valley displacement and broadened boundary transitions compared with the high-fidelity 512-view reference. This indicates that continuous sparse-view scanning degrades local structural boundaries and weakens the gray-level response of fine PCB features. Although the existing CNN-, Transformer-, and Mamba-based methods improve the overall consistency of the profile curves, residual peak attenuation, valley displacement, and boundary over-smoothing can still be observed. In contrast, RCL-Mamba produces a profile that is more consistent with the reference in terms of the main peak–valley positions and boundary transition trends, indicating improved recovery of local geometric boundaries in the via and pad regions.

Fig. 13(c) presents the profile results for the slender trace region. The LQ curve shows obvious gray-level diffusion around the trace edges, and the edge transitions are insufficiently sharp, reflecting the influence of rotational blur and sparse-view artifacts on elongated microstructures. Some comparison methods suppress global artifacts but simultaneously weaken the edge response, leading to insufficient restoration of fine details. By comparison, RCL-Mamba better preserves the boundary locations on both sides of the trace, the gray-level plateau region, and the descending edge transitions. These results demonstrate that the proposed dual-domain cascaded strategy and Mamba-CNN dual-branch module can suppress composite degradation while maintaining PCB microstructural boundaries and high-frequency details.

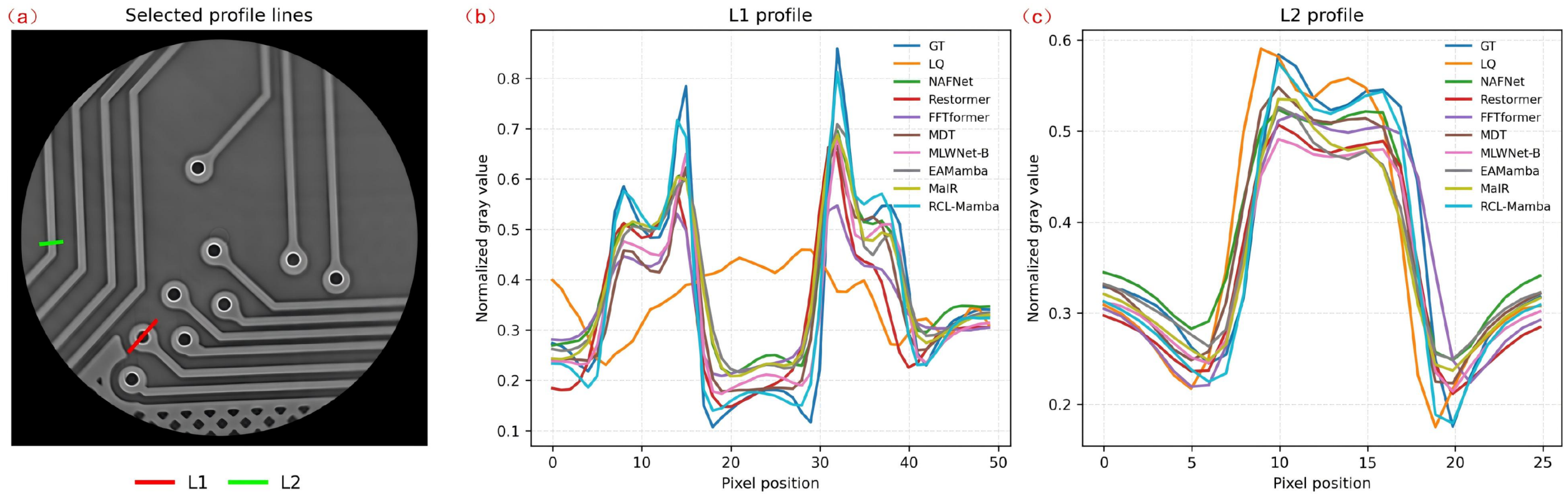

**Fig. 13.** Line-profile comparison on real-world PCB reconstructed slices. (a) Locations of the selected profile lines, where L1 crosses a via/pad region and L2 crosses a slender trace region. (b) Normalized gray-value profiles along L1. (c) Normalized gray-value profiles along L2. Here, GT denotes the high-fidelity 512-view dense-scan reconstruction reference, LQ denotes the low-quality sparse-view reconstruction, and NAFNet, Restormer, FFTformer, MDT, MLWNet-B, EAMamba, and MaIR are the comparison methods. RCL-Mamba denotes the proposed method. The reconstructed slice in (a) is displayed using a normalized intensity window of [0, 1].

To further quantify the consistency between the restored profiles and the high-fidelity 512-view reference, the root mean square error (RMSE) of each line profile was calculated, as summarized in Table 7. The RMSE was computed using the normalized gray values along each profile, with the 512-view reference profile used as the baseline. Lower RMSE values indicate better agreement with the reference and improved structural measurement fidelity. As shown in Table 7, RCL-Mamba achieves the lowest RMSE on both L1 and L2, with a mean RMSE of 0.0341, clearly outperforming the competing methods. This quantitative result is consistent with the visual profile comparison in Fig. 13, further confirming the superior capability of RCL-Mamba in preserving via/pad boundaries and slender trace structures.

**Table. 7.** Quantitative line-profile errors on real-world PCB slices.

| Methods | L1 RMSE ↓ | L2 RMSE ↓ | Mean RMSE ↓ |
|---|---|---|---|
| FDK[34] | 0.2204 | 0.0735 | 0.1470 |
| NAFNet[14] | 0.0831 | 0.0459 | 0.0645 |
| MDT[22] | 0.0836 | 0.0439 | 0.0637 |
| MLWNet-B[15] | 0.0610 | 0.0463 | 0.0537 |
| Restormer[18] | 0.1120 | 0.0464 | 0.0792 |
| FFTformer[21] | 0.1122 | 0.0412 | 0.0767 |
| EAMamba[29] | 0.0894 | 0.0491 | 0.0693 |
| MaIR[32] | 0.0886 | 0.0480 | 0.0683 |
| RCL-Mamba | 0.0407 | 0.0276 | 0.0341 |

## 5.4 Through-Hole Geometric Measurement Accuracy

In addition to image-quality metrics, accurate geometric measurement is essential for the nondestructive inspection of PCB structures. To evaluate the preservation of through-hole geometry, nine inner via apertures were selected from the reconstructed slice, as shown in Fig. 14(a). The center coordinates and diameters measured from the ground-truth reconstruction were used as the reference values. For all methods, the same regions of interest and circle-fitting procedure were employed to estimate the center coordinates and diameters of the selected apertures. The center localization error was calculated as the Euclidean distance between the fitted and reference centers, while the diameter measurement error was defined as the absolute difference between the fitted and reference diameters. According to the reconstructed voxel size, the measurement errors were converted from pixels to millimeters using a scale of 0.0255 mm/pixel.

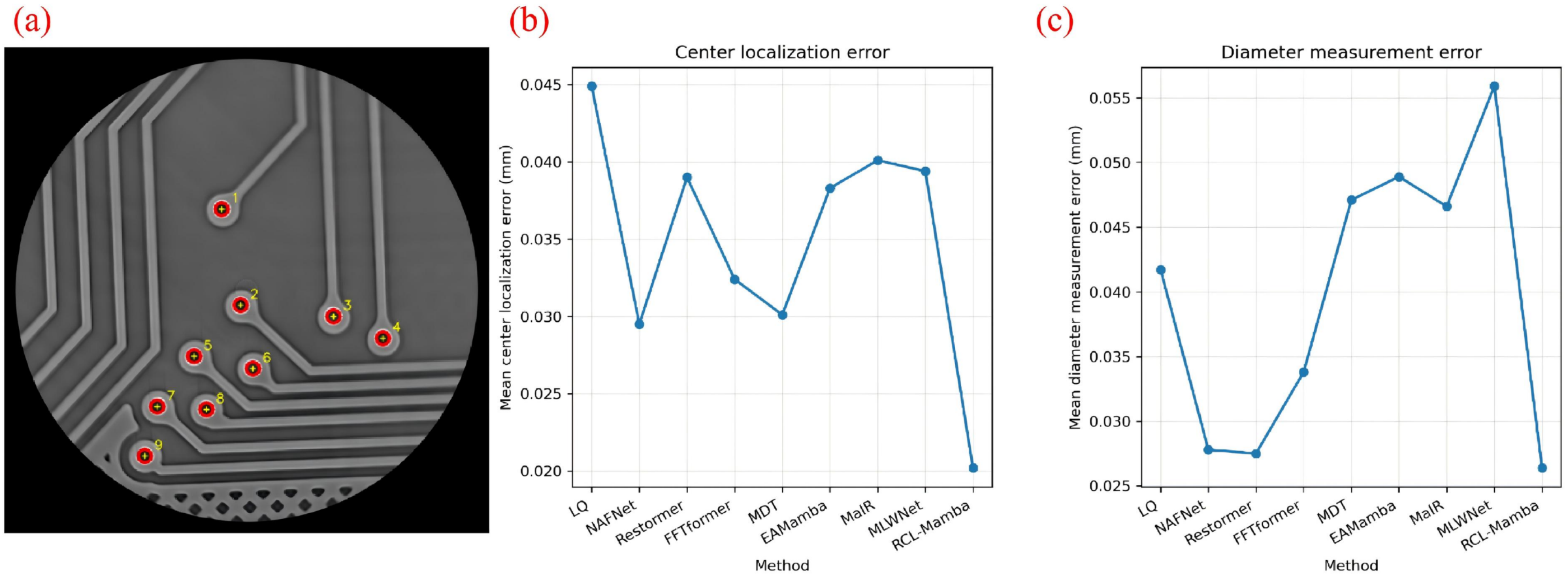

**Fig. 14.** Geometric measurement accuracy of the selected through-hole apertures. (a) Locations of the nine inner via apertures selected from the ground-truth reconstruction; (b) mean center localization errors of different methods; (c) mean diameter measurement errors of different methods. The measurement errors were calculated relative to the ground-truth reconstruction and converted to millimeters using the reconstructed voxel size of 0.0255 mm/pixel. The reconstructed slice in (a) is displayed using a normalized intensity window of [0, 1].

As shown in Fig. 14(b) and Fig. 14(c), RCL-Mamba achieves the lowest mean center localization error and diameter measurement error among all evaluated methods. Specifically, its mean center localization error is approximately 0.020 mm, compared with approximately 0.029 mm for the second-best method. Its mean diameter measurement error is approximately 0.026 mm, which is also lower than those of the comparison methods. Although NAFNet and Restormer achieve relatively low diameter measurement errors, their center localization errors remain higher than that of RCL-Mamba. These results demonstrate that the proposed dual-domain cascaded restoration strategy not only improves overall image quality but also more accurately preserves the positions and dimensions of fine through-hole structures.

### 5.5 Model Complexity

The model complexity of the proposed method was systematically compared with other advanced restoration models in terms of the number of parameters (Params), floating-point operations (FLOPs), and inference runtime, all measured at a resolution of $512\times512$ pixels. Due to space constraints, several representative methods were selected for this comparative analysis, and the quantitative results are summarized in Table 8.

**Table. 8.** Model complexity of different methods on the RCL rotational blur dataset

| Method | NAFNet | MDT | MLWNet-B | FFTformer | Restormer | EAMamba | MaIR | Ours |
|---|---|---|---|---|---|---|---|---|
| Params | 67.9M | 14.3M | 95.1M | 16.6M | 26.1M | 25.3M | 26.3M | 17.2M |
| FLOPs | 252.31G | 452.50G | 433.15G | 525.85G | 619.53G | 549.5G | 539.1G | 510.62G |
| Runtimes | 93ms | 459.3ms | 115ms | 565ms | 300ms | 234ms | 3308ms | 250ms |

As presented in Table 8, the proposed RCL-Mamba is highly lightweight, maintaining a parameter count of only 17.2 M. This represents a significant reduction in model size compared to prominent CNN-based architectures such as NAFNet (67.9 M) and MLWNet-B (95.1 M). Although the FLOPs and inference runtime of RCL-Mamba are positioned at an intermediate level, they remain highly competitive when weighed against the restoration performance. Considering the substantial quantitative and qualitative superiorities demonstrated in the previous sections, RCL-Mamba achieves a significant improvement in restoration quality within a highly reasonable computational budget. This demonstrates a favorable trade-off between reconstruction fidelity and computational complexity, further validating the practical deployment potential of the framework in high-speed industrial inspection scenarios.

### 5.6 Limitations

Although RCL-Mamba demonstrates strong performance in restoring cross-domain composite degradations, further improvement remains possible in terms of its alignment with the underlying RCL imaging physics. The current geometric transformations do not explicitly exploit the strongly anisotropic and annular characteristics of rotational blur. Future work will investigate physics-informed representations, such as polar-coordinate transformations, to represent complex arc-shaped blur trajectories in a more regular form and further improve the alignment between selective state-space scanning and the imaging physics. In addition, although the current structural measurement evaluation includes line-profile analysis and quantitative through-hole center localization and diameter measurements, the reported errors are calculated relative to the 512-view dense-scan reconstruction. Future studies will introduce calibrated standard specimens and uncertainty analysis to further evaluate absolute dimensional measurement accuracy.

## 6. Conclusion

Addressing the complex composite degradation in RCL induced by continuous exposure and sparse sampling, this paper proposes RCL-Mamba, a novel image restoration framework based on a dual-domain state space model. The framework innovatively adopts a projection-image cascaded decoupling strategy and integrates Mamba-CNN dual-branch feature modeling with a dynamic gated mechanism. This architecture effectively overcomes the inherent limitations of traditional single-domain paradigms in mitigating large-scale rotational blur and sparse artifacts. Both simulation and real-world physical experiments demonstrate that RCL-Mamba significantly outperforms existing representative algorithms in terms of global image-quality metrics, line-profile consistency, and structural measurement fidelity. Furthermore, by successfully compressing the required scanning views to 64, the proposed method yields a nearly 8-fold increase in single-station inspection efficiency. In summary, RCL-Mamba provides an advanced solution that harmonizes theoretical rigor with engineering practicality for complex restoration tasks in high-throughput industrial inspection. Moreover, it offers a novel perspective for the broader application of state space models in physically-constrained vision tasks.

**Declaration of competing interest**

The authors declare that they have no known competing financial interests or personal relationships that could have appeared to influence the work reported in this paper.

**Data availability statement**

The data that support the findings of this study are available from the corresponding author upon reasonable request.

**Acknowledgments**

This work is supported by the National Key Research and Development Program of China (No. 2022YFF0706400).

**References**

[1] Fu J, Jiang B H and Li B 2010 Large field of view computed laminography with the asymmetric rotational scanning geometry, Science China Technological Sciences 53, 2261 – 2271

[2] Tan C, Long C, Xi Y, Chen Z, Lin X, Liu F, Cai Y and Duan L 2025 Orthogonal translation computed laminography reconstruction based on self-prior information and adaptive weighted total variation, Displays 90, 103169

[3] Tan C, Wang A, Chen Z, Nong S, Liu F and Duan L 2026 Truncated projection adaptive weighting combined with adaptive TV for artifact reduction in linear computed laminography, Optics & Laser Technology 193, 114348

[4] Xue L, Hei J, Wang Y, Li Q, Lu Y and Liu W 2022 A high efficiency deep learning method for the x-ray image defect detection of casting parts, Measurement Science and Technology 33, 095015

[5] Zou X, Shi W, Du M and Xing Y 2025 Artifact reduction in rotational computed laminography using a deep learning method, Optics and Lasers in Engineering 187, 108881

[6] Long Y, Zhong Q, Lu J and Xiong C 2025 A novel detail-enhanced wavelet domain feature compensation network for sparse-view x-ray computed laminography, Journal of X-Ray Science and Technology 33, 488 – 498

[7] Liu G, Yan Y and Meng J 2024 Study on the detection technology for inner-wall outer surface defects of the automotive ABS brake master cylinder based on BM-YOLOv8, Measurement Science and Technology 35, 055109

[8] Zhao J, Zhang Y, Du X and Sun X 2024 Automated defect detection in precision forging ultrasonic images based on deep learning, Measurement Science and Technology 35, 035605

[9] Nah S, Kim T H and Lee K M 2017 Deep multi-scale convolutional neural network for dynamic scene deblurring IEEE Conf. Computer Vision and Pattern Recognition pp 3883 – 91

[10] Tao X, Gao H, Shen X, Wang J and Jia J 2018 Scale-recurrent network for deep image deblurring IEEE Conf. Computer Vision and Pattern Recognition pp 8174 – 82

[11] Zhang H, Dai Y, Li H and Koniusz P 2019 Deep stacked hierarchical multi-patch network for image deblurring IEEE/CVF Conf. Computer Vision and Pattern Recognition pp 5978 – 86

[12] Zamir S W, Arora A, Khan S, Hayat M, Khan F S, Yang M H and Shao L 2021 Multi-stage progressive image restoration IEEE/CVF Conf. Computer Vision and Pattern Recognition pp 14821 – 31

[13] Cho S J, Ji S W, Hong J P, Jung S W and Ko S J 2021 Rethinking coarse-to-fine approach in single image deblurring IEEE/CVF Int. Conf. Comput. Vis. pp 4641 – 50

[14] Chen L, Chu X, Zhang X and Sun J 2022 Simple baselines for image restoration European Conf. on Comput. Vis.

(Springer) pp 17 – 33
[15] Gao X, Qiu T, Zhang X, Bai H, Liu K, Huang X, Wei H, Zhang G and Liu H 2024 Efficient multi-scale network with learnable discrete wavelet transform for blind motion deblurring IEEE/CVF Conf. Computer Vision and Pattern Recognition pp 2733 – 42
[16] Xie W, Sun X and Ma W 2024 A light weight multi-scale feature fusion steel surface defect detection model based on YOLOv8 Meas. Sci. Technol. 35 055017
[17] Jia T, Shi L, Wei C, Shi R and Liu B 2023 Correction of motion artifact in CL based on MAFusNet J. X-ray Sci. Technol. 31 393 – 407
[18] Zamir S W, Arora A, Khan S, Hayat M, Khan F S and Yang M H 2022 Restormer: efficient transformer for high-resolution image restoration IEEE/CVF Conf. Comput. Vis. Pattern Recognit. pp 5728 – 39
[19] Tsai F J, Peng Y T, Lin Y Y, Tsai C C and Lin C W 2022 Stripformer: strip transformer for fast image deblurring European Conf. on Comput. Vis. (Springer) pp 146 – 62
[20] Wang Z, Cun X, Bao J, Zhou W, Liu J and Li H 2022 Uformer: a general u-shaped transformer for image restoration IEEE/CVF Conf. Comput. Vis. Pattern Recognit. pp 17683 – 93
[21] Kong L, Dong J, Ge J, Li M and Pan J 2023 Efficient frequency domain-based transformers for high-quality image deblurring IEEE/CVF Conf. Comput. Vis. Pattern Recognit. pp 5886 – 95
[22] Chen D, Zhou S, Pan J, Shi J, Qu L and Yang J 2025 A polarization-aided transformer for image deblurring via motion vector decomposition Comput. Vis. Pattern Recognit. Conf. pp 28061 – 70
[23] Gu A, Johnson I, Goel K, Saab K, Dao T, Rudra A and Ré C 2021 Combining recurrent, convolutional, and continuous-time models with linear state space layers Advances in Neural Information Processing Systems vol 34 pp 572 – 85
[24] Gu A, Goel K and Ré C 2021 Efficiently modeling long sequences with structured state spaces (arXiv:2111.00396)
[25] Gu A and Dao T 2024 Mamba: linear-time sequence modeling with selective state spaces First Conf. on Lang. Model.
[26] Liu Y, Tian Y, Zhao Y, Yu H, Xie L, Wang Y, Ye Q, Jiao J and Liu Y 2024 VMamba: visual state space model Advances in Neural Information Processing Systems vol 37 pp 103031 – 63
[27] Guo H, Li J, Dai T, Ouyang Z, Ren X and Xia S T 2024 MambaIR: a simple baseline for image restoration with state-space model European Conf. on Computer Vision (Springer) pp 222 – 41
[28] Shi Y, Xia B, Jin X, Wang X, Zhao T, Xia X, Xiao X and Yang W 2025 VmambaIR: visual state space model for image restoration IEEE Trans. Circuits Syst. Video Technol. 35 5560 – 74
[29] Lin Y C, Xu Y S, Chen H W, Kuo H K and Lee C Y 2025 EAMamba: efficient all-around vision state space model for image restoration IEEE/CVF Int. Conf. Comput. Vis. pp 11708 – 19
[30] Kong L, Dong J, Tang J, Yang M H and Pan J 2025 Efficient visual state space model for image deblurring Computer Vision and Pattern Recognition pp 12710 – 9
[31] Gao H, Ma B, Zhang Y, Yang J, Yang J and Dang D 2024 Learning enriched features via selective state spaces model for efficient image deblurring 32nd ACM Int. Conf. on Multimedia pp 710 – 8
[32] Li B, Zhao H, Wang W, Hu P, Gou Y and Peng X 2025 MaIR: a locality-and continuity-preserving mamba for image restoration Computer Vision and Pattern Recognition pp 7491 – 501
[33] Zhang W, Chai X, Zhu W, Zheng S, Fan G, Li Z, Zhang H and Zhang H 2023 Super-resolution reconstruction of ultrasonic Lamb wave TFM image via deep learning Meas. Sci. Technol. 34 055406
[34] Feldkamp L A, Davis L C and Kress J W 1984 Practical cone-beam algorithm J. Opt. Soc. Am. A 1 612 – 9
[35] Abbas S, Park M, Min J, Kim H K and Cho S 2014 Sparse-view computed laminography with a spherical sinusoidal scan for nondestructive testing Opt. Express 22 17745 – 55
[36] Biguri A, Dosanjh M, Hancock S and Soleimani M 2016 TIGRE: a MATLAB-GPU toolbox for CBCT image reconstruction Biomed. Phys. Eng. Express 2 055010

[37] Wang Z, Bovik A C, Sheikh H R and Simoncelli E P 2004 Image quality assessment: from error visibility to structural similarity IEEE Trans. Image Process. 13 600 – 12
[38] Xue W, Zhang L, Mou X and Bovik A C 2014 Gradient magnitude similarity deviation: a highly efficient perceptual image quality index IEEE Trans. Image Process. 23 684 – 95